\RequirePackage{fix-cm} 
\documentclass{article} 
\usepackage{iclr2027_conference,times}

\usepackage{amsmath,amsfonts,bm}

\def\eqref#1{equation~\ref{#1}}

\def\1{\bm{1}}

\DeclareMathAlphabet{\mathsfit}{\encodingdefault}{\sfdefault}{m}{sl}
\SetMathAlphabet{\mathsfit}{bold}{\encodingdefault}{\sfdefault}{bx}{n}

\usepackage{hyperref}
\usepackage{url}

\usepackage{graphicx}
\usepackage{xcolor}

\usepackage{tabularx}
\usepackage{array}
\usepackage{listings}
\usepackage{fancyvrb}

\definecolor{blue}{rgb}{0, 0, 0.8}
\definecolor{blue}{RGB}{0,85,170}

\usepackage{amsmath}    
\usepackage{amssymb}    
\usepackage{amsthm}

\usepackage{float}
\usepackage{wrapfig}

\usepackage{multirow}

\usepackage{soul}
\definecolor{darkgreen}{rgb}{0, 0.5001960, 0}
\definecolor{darkred}{rgb}{0.8, 0, 0}

\usepackage{listings}
\usepackage{enumitem}
\definecolor{diffadd}{RGB}{230,255,237}
\definecolor{diffdel}{RGB}{255,235,233}

\usepackage{soul}

\usepackage{multirow}
\usepackage{array}
\usepackage[table]{xcolor}

\usepackage{makecell}
\usepackage{array}
\usepackage{xcolor}

\usepackage{algorithm}
\usepackage{algpseudocode}
\definecolor{darkgreen}{rgb}{0,0.5,0}
\definecolor{darkred}{rgb}{0.7,0,0}

\title{AutoScientist-Quant: Self-Evolving Coding \\ Agents for Automatic Research in \\ Quantitative Investment}

\iclrfinalcopy

\author{Zongqian Li$^{1,2*}$, Yaoyiran Li$^{1\dag}$, Yaohui Guo$^{1}$, Ming Zhang$^{1}$, Nigel Collier$^{2\dag}$, Eugene Ie$^{1}$ \\
$^{1}$Google \quad $^{2}$University of Cambridge \\
\texttt{\{zqli, yaoyiran, yaohuiguo, mingzhang, eugeneie\}@google.com} \\
\texttt{\{zl510, nhc30\}@cam.ac.uk} \\
}

\begin{document}

\maketitle
\lhead{}
\chead{AutoScientist-Quant}
{\def\thefootnote{*}\footnotetext{Work done during an internship at Google.}}
{\def\thefootnote{\dag}\footnotetext{Corresponding authors.}}

\begin{abstract}
Large language model agents can discover alphas, yet current methods have three weaknesses. The search cannot adapt during the run, automation usually ends at alpha generation while library selection and model choice stay manual, and alpha discovery can read the test window through loop feedback or code problems. We present AutoScientist-Quant, a self evolving search process that regards quantitative research as one budgeted search problem. A single controller conditions every decision on the remaining budget, choosing at each round whether to improve, combine, pivot, or stop, which node to expand, how many alphas to generate, and how to retrieve past trajectories from the shared memory. The same core then selects from the library and tunes the model, closing the loop from hypothesis to deployable strategy. We also review the evaluation pipeline reused from prior work, fix two lookahead problems, and keep the feedback window disjoint from the held out test window, so every comparison tests true generalization. On CSI universes, the framework attains the best value of nearly every metric in every setting, and these conclusions hold across several backbones and markets.
\end{abstract}

\section{Introduction}
\label{Introduction}

\subsection{Background \& Related Work}
\label{sec:background}

Alphas rank stocks by expected future return \citep{fama1992cross} and are the basic input of quantitative investment \citep{bosworth1975stock, barro1990stock, demirgucc1996stock, demirgucc1996development, teweles1998stock}. Because markets are noisy and nonstationary \citep{fama1965behavior, engle1982autoregressive}, individual alphas are not effective once they become crowded, so new alphas need to be discovered at scale and renewed over time. Hand engineered libraries such as the 101 formulaic alphas \citep{kakushadze2016101} and ALPHA158 \citep{yang2020qlib} encode expert knowledge but grow slowly, while symbolic search based on reinforcement learning \citep{zhang2020autoalpha, yu2023generating} expands the formula space cheaply yet optimizes backtest fit alone, producing complex expressions that overfit and do not last.

Large language models \citep{NIPS2017_3f5ee243, minaee2025largelanguagemodelssurvey, wan2024efficient} bring financial priors and reliable code generation to this problem \citep{wu2023bloomberggptlargelanguagemodel, fu2025newquantsurveylarge, li2026breakingtrainingbottleneckseffective, li2026scalingdatadifficultyimproving}. Agentic frameworks emulate the workflow of a human researcher by proposing a market hypothesis, translating it into an alpha expression and executable code, backtesting it, and revising from feedback \citep{yao2023react, wang2023alphagpt, li2025rdagentquant}. Within this category, AlphaAgent \citep{tang2025alphaagent} regularizes the loop with originality, hypothesis consistency, and complexity bounds to keep alphas effective longer, and QuantaAlpha \citep{han2026quantaalpha} regards every end to end run as a trajectory and evolves trajectories by revising weak parts and recombining strong ones, making refinement traceable and controllable.

These methods automate alpha generation but leave the surrounding research loop fixed. Exploration follows a preset schedule with a constant number of directions, rounds, and alphas, parallel searches evolve in isolation without sharing validated experience, and the resulting library is handed to the downstream model without portfolio level selection. The gap is visible in Table~\ref{tab:alpha-results}, where the agentic baselines reach strong information quality yet negative excess returns after costs.

\subsection{Motivations}
\label{sec:motivations}

\begin{itemize}[leftmargin=*, itemsep=2pt, topsep=2pt]
\item \textbf{Static and incomplete automation in prior frameworks.} Existing designs \citep{tang2025alphaagent, han2026quantaalpha, li2025rdagentquant} fix every structural choice, predefining directions, rounds, alphas, and operations before the search starts, yet the question of which directions deserve depth is never answered. Automation also usually ends at alpha generation, handing the library to one fixed predictor with no selection or model search, leaving the model no freedom to correct earlier mistakes or explore new approaches.
\item \textbf{Train test period overlap in prior agentic pipeline.} Existing agentic pipelines \citep{tang2025alphaagent, han2026quantaalpha, li2025rdagentquant} compute loop feedback on the same window used for final evaluation. The separation of training, validation, and test data holds at the model level but not at the search level, since every accepted alpha is selected on test performance, so reported results have a selection effect toward the evaluation period \citep{bailey2017probability, harvey2015backtesting}. Overlap free evaluation requires a feedback window disjoint from the held out test window, as in Section~\ref{sec:datasets}.
\item \textbf{Test data multiple usage from prior code problem.} The evaluation code reused from prior work \citep{tang2025alphaagent, han2026quantaalpha, li2025rdagentquant} contained two lookahead problems. Information metrics were computed over the full sample rather than the declared test segment, so selection feedback incorporated the future evaluation period, and a failed segment filter scored models on their own training span. We fixed both problems and reevaluated all methods on the corrected pipeline, so the comparisons test true generalization.
\end{itemize}

\subsection{Contributions}
\label{sec:contributions}

This paper makes four contributions.

\begin{itemize}[leftmargin=*, itemsep=2pt, topsep=2pt]
\item \textbf{Resource aware dynamic search framework.} A single controller regards cost and structure as decisions rather than constants. Every choice conditions on the remaining global budget, and at each round the controller selects among \textsc{Improve}, \textsc{Combine}, \textsc{Pivot}, and \textsc{Stop}, branches from any stored node, decides how to retrieve past trajectories from the shared memory, and sets how many alphas to generate, so the search adapts to evidence while staying within a fixed spend.
\item \textbf{Closed research loop.} AutoScientist-Quant regards quantitative research as a complete scientific workflow \citep{lu2024aiscientist, aiscientist_v2, lu2026towards} rather than a single generation step. Alpha discovery, alpha filtering, and model search share one decision core, every round runs the full cycle of review, ideation, experiment, analysis, and reflection, and each experiment is itself a complete development pass through coding and debugging, training, and backtesting, with model selection and parameter tuning in the model search stage.
\item \textbf{Trustworthy evaluation infrastructure.} We reviewed the evaluation code reused from prior work, corrected two lookahead problems, and separated the feedback window from the held out test window at the design level, then reevaluated every method on the corrected pipeline. Thus, all reported comparisons rest on an overlap free foundation.
\item \textbf{Validated alpha library and financial insights.} The discovered alphas earn positive excess returns after costs on the held out test window. Each alpha comes with a readable history, the hypothesis that produced it, the revisions that refined it, and the evidence that kept it in the library. The process yields transferable insights on how alphas are found, how they are improved, and what the selected ones reveal about the market.
\end{itemize}

\section{Methods}
\label{Methods}

\begin{figure}[t!]
\centering
\includegraphics[width=0.7\textwidth]{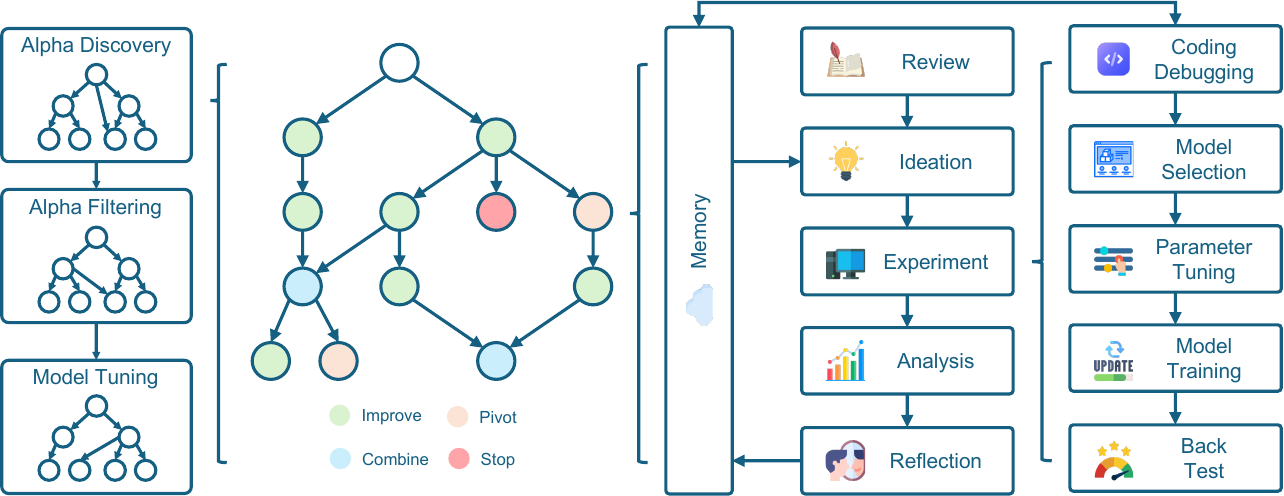}
\caption{\textbf{Overview of AutoScientist-Quant.} Left: the three stages, each a budgeted tree search. Middle: the search, where every node is a round of a trajectory, colors mark the options chosen by the controller, and all trajectories share one memory. Right: the agent cycle inside each round and the pipeline inside each experiment.}
\vspace{-2em}
\label{fig:method}
\end{figure}

AutoScientist-Quant regards quantitative research as one budgeted search problem solved in three linked stages, overviewed in Figure~\ref{fig:method}. Alpha discovery generates a candidate alpha library, alpha filtering selects it, and model search tunes the predictor that consumes it. All three stages share one decision core. At every round a controller conditions on the remaining global budget and chooses the option among \textsc{Improve}, \textsc{Combine}, \textsc{Pivot}, and \textsc{Stop}, the parent node to expand, and the number of candidates to generate. The same process is reused on the alpha space, the library subset space, and the model configuration space.

%

\subsection{Problem Formulation}
\label{sec:formulation}

Let $\mathbf{X}_{\le t}\in\mathbb{R}^{N\times t\times W}$ collect the history of $N$ stocks up to day $t$, with $W$ price and volume fields per stock and day. An alpha is a function $f$ mapping this history to cross sectional scores $f(\mathbf{X}_{\le t})\in\mathbb{R}^{N}$, a library $\mathcal{F}=\{f_1,\dots,f_K\}$ combines the scores into inputs, and a predictor $g\in\mathcal{G}$ maps inputs to forecasts $\hat{\mathbf{y}}_t=g\!\left(\mathcal{F}(\mathbf{X}_{\le t})\right)\in\mathbb{R}^{N}$ of the future return. A fixed portfolio rule turns forecasts into daily positions, and $U(\mathcal{F},g;\mathcal{D})$ denotes the combined backtest utility on window $\mathcal{D}$, aggregating the metrics of Section~\ref{sec:metrics}, either through the controller's own reading of these metrics during alpha discovery or through the exact standardized score of Eq.~\ref{eq:zscore} during filtering and model search. The framework solves
\begin{equation}
\label{eq:objective}
\max_{\mathcal{F}_B,\ \mathcal{F}\subseteq\mathcal{F}_B,\ g\in\mathcal{G}}\ U\!\left(\mathcal{F}\cup\mathcal{A},\,g;\ \mathcal{D}_{\mathrm{fb}}\right)
\qquad\text{s.t.}\qquad |\mathcal{F}_B|\le B,
\end{equation}
where $\mathcal{F}_B$ ranges over libraries discoverable under the global budget $B$, $\mathcal{A}$ is an optional standard library, and $\mathcal{D}_{\mathrm{fb}}$ is the feedback window of Section~\ref{sec:datasets}. The held out window enters exactly once, to produce the numbers in Section~\ref{Results}.

\subsection{Alpha Discovery}
\label{sec:discovery}

Alpha discovery runs trajectories. A trajectory is a sequence $\tau=\left((h_1,\mathcal{F}_1,e_1),\dots,(h_R,\mathcal{F}_R,e_R)\right)$ whose round $r$ proposes a market hypothesis $h_r$, builds and executes the alpha set $\mathcal{F}_r$ implementing it, and records backtest feedback $e_r$; its reward is $R(\tau)=\max_r V(\mathcal{F}_r)$ with $V$ the feedback window utility. All trajectories from all directions live in a shared pool $\mathcal{P}$, the working memory, visible to every branch. At round $r$ the controller observes the pool and the remaining budget $b_r$ and outputs one decision per active branch $d$,
\begin{equation}
\label{eq:controller}
(o_{d,r},\sigma_{d,r},n_{d,r})=\pi_{\mathrm{LLM}}\!\left(\mathcal{P}_r,\,b_r\right),
\qquad
o_{d,r}\in\{\textsc{Improve},\textsc{Combine},\textsc{Pivot},\textsc{Stop}\},
\end{equation}
where the position $\sigma_{d,r}$ selects parent nodes in $\mathcal{P}_r$ and the cardinality $n_{d,r}\in[n_{\min},n_{\max}]$ caps how many alphas the round may generate. \textsc{Improve} performs a targeted revision, localizing a step of the parent trajectory by self reflection and rewriting only that part, and \textsc{Combine} recombines complementary parts of stored parents, where the strategy for retrieving parents from the pool is itself decided by the controller at run time; the two options follow \citet{han2026quantaalpha}. \textsc{Pivot} opens a fresh direction from a new market hypothesis when existing directions plateau, and \textsc{Stop} closes a branch whose expected gain no longer justifies its cost. The budget evolves as
\begin{equation}
\label{eq:budget}
b_{r+1}=b_r-\sum\nolimits_{d} n_{d,r},
\qquad
\pi_{\mathrm{LLM}}(\cdot\mid b=0)=\textsc{Stop},
\end{equation}
so the controller allocates search capacity where marginal evidence is strongest and stops once it is not obvious.

\subsection{Alpha Filtering}
\label{sec:filtering}

Discovery returns a library whose members are individually validated but jointly redundant. The library seeds a subset search guided by the same controller. The state is a split of the library into a selected pool $S$ and a removed pool; the edits are remove, add, and swap; and \textsc{Improve}, \textsc{Combine}, and \textsc{Pivot} extend, recombine, or restart edit sequences from any stored node. Each candidate subset is scored against the historic pool of nodes,
\begin{equation}
\label{eq:zscore}
Z(S)=\sum_{k}\frac{m_{k}(S)-\mu_{k}}{\varsigma_{k}},
\end{equation}
where $m_k(S)$ is the value of metric $k$ for node $S$, and $\mu_k,\varsigma_k$ standardize metric $k$ against its historic distribution across previously evaluated nodes. The sum covers the eight metrics of Section~\ref{sec:metrics}, and all nodes are rescored against the current pool at each comparison, so every ranking uses one set of statistics. Both this stage and the model search stage reuse the budget rule of Eq.~\ref{eq:budget}, with $n_{d,r}$ counting evaluated candidates, subset edits here and model configurations in Section~\ref{sec:model-searching}.

\subsection{Model Search}
\label{sec:model-searching}

The final stage reuses the controller on the model configuration space $\Omega=\bigcup_{m\in\mathcal{M}}\{m\}\times\Theta_{m}$, where $\mathcal{M}$ contains Linear, LightGBM \citep{ke2017lightgbm}, XGBoost \citep{chen2016xgboost}, CatBoost \citep{prokhorenkova2018catboost}, and other models and $\Theta_{m}$ its hyperparameter space, so a candidate is a complete algorithm and hyperparameter choice. Candidates branch from any earlier configuration through \textsc{Improve}, \textsc{Combine}, and \textsc{Pivot} and are scored by the standardized utility of Eq.~\ref{eq:zscore} with $S$ replaced by $(m,\theta_{m})$. The winning pair of library and configuration is trained once on the training and validation windows of Section~\ref{sec:datasets}, with early stopping on validation, and evaluated once on the held out window, which no stage of the search ever observes.

\section{Experimental Design}
\label{Experimental Design}

\subsection{Datasets}
\label{sec:datasets}

We use daily open, high, low, close, and volume data of six universes, \textbf{CSI 300}, \textbf{CSI 500}, \textbf{CSI 800}, \textbf{CSI 1000}, \textbf{S\&P 500}, and \textbf{NASDAQ 100}, from January 2015 to May 2026, accessed through Qlib \citep{yang2020qlib} with index members taken as of each historical date. The main experiments in Table~\ref{tab:alpha-results} use CSI 300, the market robustness analysis in Table~\ref{tab:ablation-market} covers the other three CSI universes, and Table~\ref{tab:us-markets} covers the two US universes. The data are split into four continuous windows, training from January 2015 to January 2020, validation from January 2020 to June 2021, a feedback window from June 2021 to June 2023 that supplies backtest feedback to the agents, and a test window from June 2023 to May 2026 reserved for final evaluation. The test window never overlaps any window the agents observe, so information from the evaluation period cannot reach the discovery loop.
\subsection{Evaluation Metrics}
\label{sec:metrics}

We report four information metrics and four portfolio metrics, with each result averaged over three independent runs. The information coefficient \textbf{IC} is the Pearson correlation between predicted scores and realized labels within each daily cross section, \textbf{ICIR} is the mean daily IC over its standard deviation, and \textbf{RIC} (Rank IC) and \textbf{RICIR} (Rank ICIR) are the Spearman counterparts. Portfolio metrics come from a cost adjusted daily strategy that holds the 50 stocks with the highest scores, replaces at most 5 of them per day, and is benchmarked against the index of each universe. \textbf{ARR} is the annualized excess return over the benchmark index net of transaction costs, \textbf{IR} is the information ratio of daily excess returns, \textbf{MDD} is the maximum decrease of the cumulative excess return, and \textbf{CR} is the Calmar ratio equal to ARR divided by the absolute value of MDD. On the US universes these metrics are computed on absolute returns without subtracting the index return, as noted in Table~\ref{tab:us-markets}.

\subsection{Baselines}
\label{sec:baselines}

Baselines span three categories over three standard Qlib alpha libraries, ALPHA158, ALPHA360, and ALPHA20 \citep{yang2020qlib}. Machine learning baselines include \textbf{Linear}, \textbf{LightGBM} \citep{ke2017lightgbm}, and \textbf{CatBoost} \citep{prokhorenkova2018catboost}, and deep learning baselines include \textbf{LSTM} \citep{hochreiter1997long}, \textbf{TCN} \citep{bai2018empirical}, and \textbf{KRNN} \citep{CHATTHA2025126148}. The agentic baselines \textbf{AlphaAgent} \citep{tang2025alphaagent} and \textbf{QuantaAlpha} \citep{han2026quantaalpha} of Section~\ref{sec:background} share our backbone, search budget, and backtest procedure. R\&D-Agent-Quant \citep{li2025rdagentquant} is excluded since its loop selects alphas on the window it reports, the overlap of Section~\ref{sec:motivations}. Every resulting library is backtested with a fixed LightGBM predictor before the model search stage, evaluated alone as CUSTOM, and merged with each standard library as the +ALPHA settings.

\subsection{Models}
\label{sec:models}

All agentic methods use the same large language model backbone within each comparison. We run two backbones, \textbf{GPT-OSS-120B} \citep{openai2025gptoss} and \textbf{GLM-5} \citep{glm5team2026glm5vibecodingagentic}, referred to as GPT and GLM in the tables. The main results and ablations in Tables~\ref{tab:alpha-results} and \ref{tab:ablation-no-modelsearch} use GPT, the model robustness analysis in Table~\ref{tab:method-comparison} repeats the comparison under GLM, and the market robustness analysis in Table~\ref{tab:ablation-market} runs both backbones on three further universes, so every conclusion is checked against a change of backbone. The framework itself is model agnostic and also supports other backbones.

\begin{table}[t!]
\centering
\setlength{\tabcolsep}{7.5pt}
\fontsize{6}{8}\selectfont
\begin{tabular}{l l c c c c c c c c}
\hline
\textbf{Method} & \textbf{Alpha} & \textbf{IC} & \textbf{ICIR} & \textbf{RIC} & \textbf{RICIR} & \textbf{ARR} & \textbf{IR} & \textbf{MDD} & \textbf{CR} \\
\hline
\multicolumn{10}{c}{\textbf{\textit{Machine Learning Methods}}} \\
\hline
\multirow{3}{*}{LightGBM}
 & ALPHA20  & -0.003 & -0.019 & 0.014 & 0.079 & -1.0\% & -0.100 & -24.9\% & -0.039 \\
 & ALPHA158 & 0.010  & 0.063  & 0.029 & 0.176 & -0.1\% & -0.007 & -13.2\% & -0.004 \\
 & ALPHA360 & 0.006  & 0.036  & 0.025 & 0.147 & -0.1\% & -0.007 & -17.3\% & -0.003 \\
\hline
\multirow{3}{*}{Linear}
 & ALPHA20  & -0.006 & -0.033 & 0.010 & 0.051 & -7.1\% & -0.686 & -31.6\% & -0.224 \\
 & ALPHA158 & 0.005  & 0.025  & 0.020 & 0.102 & -5.1\% & -0.562 & -20.1\% & -0.252 \\
 & ALPHA360 & -0.002 & -0.013 & 0.014 & 0.072 & -6.5\% & -0.665 & -23.9\% & -0.272 \\
\hline
\multirow{3}{*}{CatBoost}
 & ALPHA20  & -0.003 & -0.013 & 0.012 & 0.062 & -2.7\% & -0.259 & -26.3\% & -0.102 \\
 & ALPHA158 & 0.009  & 0.050  & 0.027 & 0.157 & -2.1\% & -0.275 & -13.7\% & -0.153 \\
 & ALPHA360 & 0.008  & 0.043  & 0.025 & 0.141 & -0.0\% & -0.004 & -16.1\% & -0.002 \\
\hline
\multicolumn{10}{c}{\textbf{\textit{Deep Learning Methods}}} \\
\hline
\multirow{3}{*}{LSTM}
 & ALPHA20  & -0.009 & -0.042 & 0.008 & 0.035 & -9.9\% & -0.916 & -32.2\% & -0.307 \\
 & ALPHA158 & 0.006  & 0.043  & 0.017 & 0.151 & 0.5\%  & 0.092  & -7.5\% & 0.072 \\
 & ALPHA360 & -0.000 & -0.001 & 0.021 & 0.110 & -4.5\% & -0.552 & -25.3\% & -0.177 \\
\hline
\multirow{3}{*}{TCN}
 & ALPHA20  & -0.006 & -0.033 & 0.007 & 0.035 & -3.7\% & -0.347 & -30.8\% & -0.121 \\
 & ALPHA158 & 0.007  & 0.039  & 0.026 & 0.152 & -1.5\% & -0.190 & -12.9\% & -0.114 \\
 & ALPHA360 & -0.000 & -0.002 & 0.013 & 0.068 & -0.3\% & -0.027 & -20.4\% & -0.013 \\
\hline
\multirow{3}{*}{KRNN}
 & ALPHA20  & 0.001  & 0.008  & 0.014 & 0.097 & -2.5\% & -0.280 & -20.3\% & -0.123 \\
 & ALPHA158 & 0.003  & 0.025  & 0.018 & 0.150 & -4.2\% & -0.723 & -14.0\% & -0.296 \\
 & ALPHA360 & 0.003  & 0.017  & 0.025 & 0.152 & -4.8\% & -0.673 & -21.6\% & -0.221 \\
\hline
\multicolumn{10}{c}{\textbf{\textit{Agentic Methods}}} \\
\hline
\multirow{4}{*}{AlphaAgent}
 & CUSTOM    & 0.024 & 0.147 & 0.022 & 0.135 & -2.3\% & -0.321 & -15.1\% & -0.153 \\
 & +ALPHA20  & 0.023 & 0.135 & 0.020 & 0.121 & -1.0\% & -0.130 & -17.0\% & -0.055 \\
 & +ALPHA158 & 0.030 & 0.182 & 0.027 & 0.168 & -1.0\% & -0.151 & -12.5\% & -0.087 \\
 & +ALPHA360 & 0.027 & 0.164 & 0.026 & 0.155 & -1.5\% & -0.202 & -15.5\% & -0.098 \\
\hline
\multirow{4}{*}{QuantaAlpha}
 & CUSTOM    & 0.024 & 0.144 & 0.022 & 0.130 & -1.6\% & -0.207 & -14.5\% & -0.096 \\
 & +ALPHA20  & 0.023 & 0.139 & 0.021 & 0.125 & -0.5\% & -0.060 & -15.6\% & -0.033 \\
 & +ALPHA158 & 0.031 & 0.190 & 0.029 & 0.177 & -0.2\% & -0.018 & -12.5\% & -0.013 \\
 & +ALPHA360 & 0.029 & 0.173 & 0.027 & 0.160 & -1.5\% & -0.189 & -13.3\% & -0.098 \\
\hline
\multirow{4}{*}{\shortstack[l]{AutoScientist\\-Quant}}
 & CUSTOM    & \textbf{0.028} & \textbf{0.185} & \textbf{0.026} & \textbf{0.172} & \textbf{1.8\%} & \textbf{0.259} & \textbf{-12.7\%} & \textbf{0.143} \\
 & +ALPHA20  & \textbf{0.028} & \textbf{0.181} & \textbf{0.026} & \textbf{0.165} & \textbf{2.4\%} & \textbf{0.330} & \textbf{-12.5\%} & \textbf{0.209} \\
 & +ALPHA158 & \textbf{0.034} & \textbf{0.219} & \textbf{0.032} & \textbf{0.206} & \textbf{3.5\%} & \textbf{0.500} & \textbf{-9.5\%} & \textbf{0.368} \\
 & +ALPHA360 & \textbf{0.033} & \textbf{0.212} & \textbf{0.031} & \textbf{0.198} & \textbf{3.0\%} & \textbf{0.401} & \textbf{-12.0\%} & \textbf{0.263} \\
\hline
\end{tabular}
\caption{\textbf{Main results on CSI 300.} Machine learning, deep learning, and agentic methods across four alpha settings, each library used alone and merged with a standard library. Agentic methods use the GPT backbone. ARR is the annualized excess return over the CSI 300 index, net of transaction costs. AutoScientist-Quant reports the full three stage framework, and the ablation study in Table~\ref{tab:ablation-no-modelsearch} shows that alpha discovery alone, without alpha filtering or model search, still outperforms every baseline on almost every metric.}
\vspace{-2em}
\label{tab:alpha-results}
\end{table}

\section{Results}
\label{Results}

\subsection{Main Results}
\label{sec:main-results}

\newcommand{\negv}[1]{\textcolor{darkred}{#1}}
\newcommand{\posv}[1]{\textcolor{darkgreen}{#1}}

\newcommand{\tcell}[3]{%
  \setlength{\arraycolsep}{0pt}%
  $\vcenter{\hbox{#2}}\,
   \vcenter{\hbox{$\begin{array}{@{}l@{}}
     \raisebox{0.6ex}{\tiny#1}\\[-2.8ex]
     \raisebox{-0.3ex}{\tiny#3}
   \end{array}$}}$}

\newcommand{\blockrowsep}{-2pt}

\begin{table}[t!]
\centering
\setlength{\tabcolsep}{6pt}
\renewcommand{\arraystretch}{1.2}
\fontsize{8}{10}\selectfont
\resizebox{0.85\textwidth}{!}{%
\begin{tabular}{l c c c c c c c c}
\hline
\textbf{Alpha} & \textbf{IC} & \textbf{ICIR} & \textbf{RIC} & \textbf{RICIR} & \textbf{ARR} & \textbf{IR} & \textbf{MDD} & \textbf{CR} \\
\hline
\multicolumn{9}{c}{\textbf{\textit{AutoScientist-Quant (w/o Model Search) = Alpha Discovery + Alpha Filtering}}} \\
\hline
CUSTOM    & \tcell{\posv{+0.002}}{0.026}{\negv{-0.002}} & \tcell{\posv{+0.008}}{0.155}{\negv{-0.030}} & \tcell{\posv{+0.003}}{0.025}{\negv{-0.002}} & \tcell{\posv{+0.014}}{0.148}{\negv{-0.024}} & \tcell{\posv{+2.5\%}}{0.9\%}{\negv{-0.9\%}} & \tcell{\posv{+0.337}}{0.130}{\negv{-0.129}} & \tcell{\posv{+1.2\%}}{-13.3\%}{\negv{-0.6\%}} & \tcell{\posv{+0.162}}{0.066}{\negv{-0.077}} \\
+ALPHA20  & \tcell{\posv{+0.003}}{0.026}{\negv{-0.002}} & \tcell{\posv{+0.017}}{0.156}{\negv{-0.025}} & \tcell{\posv{+0.004}}{0.025}{\negv{-0.001}} & \tcell{\posv{+0.024}}{0.148}{\negv{-0.016}} & \tcell{\posv{+2.6\%}}{2.1\%}{\negv{-0.3\%}} & \tcell{\posv{+0.354}}{0.295}{\negv{-0.035}} & \tcell{\posv{+2.0\%}}{-13.6\%}{\negv{-1.2\%}} & \tcell{\posv{+0.226}}{0.194}{\negv{-0.016}} \\
+ALPHA158 & \tcell{\posv{+0.002}}{0.033}{\negv{-0.001}} & \tcell{\posv{+0.008}}{0.198}{\negv{-0.021}} & \tcell{\posv{+0.002}}{0.031}{\negv{-0.001}} & \tcell{\posv{+0.012}}{0.189}{\negv{-0.017}} & \tcell{\posv{+1.7\%}}{1.5\%}{\negv{-2.0\%}} & \tcell{\posv{+0.239}}{0.221}{\negv{-0.279}} & \tcell{\posv{+2.6\%}}{-9.9\%}{\negv{-0.4\%}} & \tcell{\posv{+0.179}}{0.167}{\negv{-0.201}} \\
+ALPHA360 & \tcell{\posv{+0.003}}{0.032}{\negv{-0.001}} & \tcell{\posv{+0.014}}{0.187}{\negv{-0.025}} & \tcell{\posv{+0.004}}{0.030}{\negv{-0.001}} & \tcell{\posv{+0.020}}{0.179}{\negv{-0.019}} & \tcell{\posv{+2.3\%}}{0.8\%}{\negv{-2.2\%}} & \tcell{\posv{+0.315}}{0.125}{\negv{-0.276}} & \tcell{\posv{+0.5\%}}{-12.9\%}{\negv{-0.9\%}} & \tcell{\posv{+0.185}}{0.087}{\negv{-0.176}} \\
\hline
\multicolumn{9}{c}{\textbf{\textit{AutoScientist-Quant (w/o Alpha Filtering or Model Search) = Alpha Discovery}}} \\
\hline
CUSTOM    & \tcell{\posv{+0.001}}{0.026}{\negv{-0.003}} & \tcell{\posv{+0.005}}{0.153}{\negv{-0.033}} & \tcell{\posv{+0.003}}{0.024}{\negv{-0.002}} & \tcell{\posv{+0.012}}{0.147}{\negv{-0.025}} & \tcell{\posv{+1.3\%}}{-0.3\%}{\negv{-2.1\%}} & \tcell{\posv{+0.184}}{-0.022}{\negv{-0.281}} & \tcell{\negv{-0.6\%}}{-15.1\%}{\negv{-2.4\%}} & \tcell{\posv{+0.096}}{0.000}{\negv{-0.143}} \\
+ALPHA20  & \tcell{\posv{+0.002}}{0.026}{\negv{-0.003}} & \tcell{\posv{+0.013}}{0.152}{\negv{-0.029}} & \tcell{\posv{+0.003}}{0.024}{\negv{-0.002}} & \tcell{\posv{+0.020}}{0.145}{\negv{-0.020}} & \tcell{\posv{+1.7\%}}{1.2\%}{\negv{-1.2\%}} & \tcell{\posv{+0.233}}{0.174}{\negv{-0.156}} & \tcell{\posv{+1.4\%}}{-14.2\%}{\negv{-1.8\%}} & \tcell{\posv{+0.160}}{0.127}{\negv{-0.082}} \\
+ALPHA158 & \tcell{\posv{+0.001}}{0.033}{\negv{-0.002}} & \tcell{\posv{+0.005}}{0.195}{\negv{-0.025}} & \tcell{\posv{+0.002}}{0.031}{\negv{-0.001}} & \tcell{\posv{+0.008}}{0.185}{\negv{-0.021}} & \tcell{\posv{+0.7\%}}{0.6\%}{\negv{-3.0\%}} & \tcell{\posv{+0.097}}{0.079}{\negv{-0.422}} & \tcell{\posv{+2.5\%}}{-10.0\%}{\negv{-0.5\%}} & \tcell{\posv{+0.077}}{0.064}{\negv{-0.304}} \\
+ALPHA360 & \tcell{\posv{+0.002}}{0.031}{\negv{-0.002}} & \tcell{\posv{+0.009}}{0.182}{\negv{-0.030}} & \tcell{\posv{+0.003}}{0.029}{\negv{-0.002}} & \tcell{\posv{+0.014}}{0.174}{\negv{-0.024}} & \tcell{\posv{+1.3\%}}{-0.1\%}{\negv{-3.1\%}} & \tcell{\posv{+0.194}}{0.005}{\negv{-0.397}} & \tcell{\negv{-0.9\%}}{-14.2\%}{\negv{-2.2\%}} & \tcell{\posv{+0.138}}{0.040}{\negv{-0.223}} \\
\hline
\multicolumn{9}{c}{\textbf{\textit{Alpha Discovery (w/o Searching Strategy)}}} \\
\hline
CUSTOM    & \tcell{\phantom{-0.000}}{0.025}{\negv{-0.003}} & \tcell{\phantom{-0.000}}{0.145}{\negv{-0.040}} & \tcell{\phantom{-0.000}}{0.023}{\negv{-0.004}} & \tcell{\phantom{-0.000}}{0.133}{\negv{-0.040}} & \tcell{\phantom{-0.0\%}}{-2.4\%}{\negv{-4.1\%}} & \tcell{\phantom{-0.000}}{-0.291}{\negv{-0.551}} & \tcell{\phantom{-0.0\%}}{-16.5\%}{\negv{-3.8\%}} & \tcell{\phantom{-0.000}}{-0.126}{\negv{-0.269}} \\[\blockrowsep]
+ALPHA20  & \tcell{\phantom{-0.000}}{0.024}{\negv{-0.005}} & \tcell{\phantom{-0.000}}{0.138}{\negv{-0.043}} & \tcell{\phantom{-0.000}}{0.021}{\negv{-0.005}} & \tcell{\phantom{-0.000}}{0.125}{\negv{-0.039}} & \tcell{\phantom{-0.0\%}}{-0.7\%}{\negv{-3.1\%}} & \tcell{\phantom{-0.000}}{-0.086}{\negv{-0.416}} & \tcell{\phantom{-0.0\%}}{-15.0\%}{\negv{-2.6\%}} & \tcell{\phantom{-0.000}}{-0.034}{\negv{-0.244}} \\[\blockrowsep]
+ALPHA158 & \tcell{\phantom{-0.000}}{0.031}{\negv{-0.003}} & \tcell{\phantom{-0.000}}{0.189}{\negv{-0.030}} & \tcell{\phantom{-0.000}}{0.029}{\negv{-0.003}} & \tcell{\phantom{-0.000}}{0.178}{\negv{-0.028}} & \tcell{\phantom{-0.0\%}}{-0.1\%}{\negv{-3.7\%}} & \tcell{\phantom{-0.000}}{-0.022}{\negv{-0.522}} & \tcell{\phantom{-0.0\%}}{-12.0\%}{\negv{-2.5\%}} & \tcell{\phantom{-0.000}}{-0.008}{\negv{-0.375}} \\[\blockrowsep]
+ALPHA360 & \tcell{\phantom{-0.000}}{0.030}{\negv{-0.003}} & \tcell{\phantom{-0.000}}{0.182}{\negv{-0.030}} & \tcell{\phantom{-0.000}}{0.028}{\negv{-0.003}} & \tcell{\phantom{-0.000}}{0.173}{\negv{-0.026}} & \tcell{\phantom{-0.0\%}}{-0.4\%}{\negv{-3.4\%}} & \tcell{\phantom{-0.000}}{-0.051}{\negv{-0.452}} & \tcell{\phantom{-0.0\%}}{-13.5\%}{\negv{-1.6\%}} & \tcell{\phantom{-0.000}}{-0.030}{\negv{-0.293}} \\
\hline
\multicolumn{9}{c}{\textbf{\textit{Alpha Discovery (w/o Dynamics)}}} \\
\hline
CUSTOM    & \tcell{\phantom{-0.000}}{0.024}{\negv{-0.004}} & \tcell{\phantom{-0.000}}{0.142}{\negv{-0.043}} & \tcell{\phantom{-0.000}}{0.022}{\negv{-0.004}} & \tcell{\phantom{-0.000}}{0.132}{\negv{-0.040}} & \tcell{\phantom{-0.0\%}}{-2.9\%}{\negv{-4.6\%}} & \tcell{\phantom{-0.000}}{-0.389}{\negv{-0.648}} & \tcell{\phantom{-0.0\%}}{-16.9\%}{\negv{-4.2\%}} & \tcell{\phantom{-0.000}}{-0.152}{\negv{-0.295}} \\[\blockrowsep]
+ALPHA20  & \tcell{\phantom{-0.000}}{0.025}{\negv{-0.003}} & \tcell{\phantom{-0.000}}{0.153}{\negv{-0.028}} & \tcell{\phantom{-0.000}}{0.023}{\negv{-0.002}} & \tcell{\phantom{-0.000}}{0.143}{\negv{-0.021}} & \tcell{\phantom{-0.0\%}}{-2.6\%}{\negv{-5.0\%}} & \tcell{\phantom{-0.000}}{-0.363}{\negv{-0.693}} & \tcell{\phantom{-0.0\%}}{-14.4\%}{\negv{-1.9\%}} & \tcell{\phantom{-0.000}}{-0.168}{\negv{-0.378}} \\[\blockrowsep]
+ALPHA158 & \tcell{\phantom{-0.000}}{0.032}{\negv{-0.002}} & \tcell{\phantom{-0.000}}{0.189}{\negv{-0.030}} & \tcell{\phantom{-0.000}}{0.029}{\negv{-0.002}} & \tcell{\phantom{-0.000}}{0.177}{\negv{-0.030}} & \tcell{\phantom{-0.0\%}}{0.5\%}{\negv{-3.0\%}} & \tcell{\phantom{-0.000}}{0.078}{\negv{-0.422}} & \tcell{\phantom{-0.0\%}}{-11.2\%}{\negv{-1.7\%}} & \tcell{\phantom{-0.000}}{0.045}{\negv{-0.323}} \\[\blockrowsep]
+ALPHA360 & \tcell{\phantom{-0.000}}{0.028}{\negv{-0.005}} & \tcell{\phantom{-0.000}}{0.170}{\negv{-0.042}} & \tcell{\phantom{-0.000}}{0.026}{\negv{-0.005}} & \tcell{\phantom{-0.000}}{0.159}{\negv{-0.039}} & \tcell{\phantom{-0.0\%}}{-1.2\%}{\negv{-4.2\%}} & \tcell{\phantom{-0.000}}{-0.167}{\negv{-0.569}} & \tcell{\phantom{-0.0\%}}{-13.2\%}{\negv{-1.3\%}} & \tcell{\phantom{-0.000}}{-0.092}{\negv{-0.355}} \\
\hline
\end{tabular}%
}
\caption{\textbf{Ablation of AutoScientist-Quant on CSI 300 under GPT.} The center value in each cell is the absolute metric, the top right corner, shown for the first two parts, is the gain over the best agentic value per metric, and the bottom right corner is the gap to our full method, both computed from unrounded values. \textcolor{darkgreen}{Green} marks positive values and \textcolor{darkred}{red} negative.}
\vspace{-0em}
\label{tab:ablation-no-modelsearch}
\end{table}

Table~\ref{tab:alpha-results} compares the three categories on CSI 300 under the GPT backbone.

\textbf{Agentic generation creates alpha beyond fixed input sets.} The best machine learning or deep learning result reaches an IC of 0.010, obtained by LightGBM on ALPHA158, and most of these strategies lose to the index. All three agentic methods lift IC to between 0.023 and 0.034, roughly three times the best static result, so hypothesis guided discovery adds information beyond predefined inputs.

\textbf{AutoScientist-Quant is the strongest agentic method in every setting.} It attains the best value of all eight metrics in all four alpha settings among agentic methods. With ALPHA158 it reaches an IC of 0.034 and an ICIR of 0.219, against 0.031 and 0.190 for QuantaAlpha and 0.030 and 0.182 for AlphaAgent, and the same ordering holds for the rank based metrics.

\textbf{AutoScientist-Quant turns alpha into strong portfolio value.} Its excess return is positive in all four settings, rising from 1.8\% with discovered alphas alone to 3.5\% with ALPHA158, where the information ratio reaches 0.500 and the Calmar ratio 0.368. QuantaAlpha with ALPHA158 pairs an IC of 0.031 with an ARR of $-0.2\%$, and among agentic methods AutoScientist-Quant also keeps the smallest MDD in every setting. This is the gap anticipated in Section~\ref{sec:motivations}. Information metrics alone do not price turnover, costs, and tail behavior.

\textbf{Generated and predefined alphas are complements.} For every agentic method the merged settings outperform its own library alone, with ALPHA158 the strongest partner for all three methods, and AutoScientist-Quant rises from an IC of 0.028 and an ARR of 1.8\% with CUSTOM to 0.034 and 3.5\% with ALPHA158. The complement runs both ways, since even our CUSTOM setting, which uses no predefined alphas, earns more than any baseline setting that includes them.

\subsection{Ablation Studies}
\label{sec:ablation}

Table~\ref{tab:ablation-no-modelsearch} removes the stages of AutoScientist-Quant one at a time on CSI 300 under GPT. The first part keeps alpha discovery and alpha filtering and removes model search, the second part keeps alpha discovery alone, and the last two parts remove single components inside alpha discovery. The searching strategy denotes the \textsc{Improve}, \textsc{Combine}, \textsc{Pivot}, and \textsc{Stop} options of Section~\ref{sec:discovery}, and the dynamics denotes the per round decisions on alpha count, early stopping, and direction creation. A removed component follows the setting of the agentic baselines.

\textbf{The downstream stages turn alpha into returns.} Removing model search leaves IC nearly unchanged yet costs up to 2.2 percentage points of excess return, and removing alpha filtering as well pushes the CUSTOM setting from an ARR of 0.9\% to $-0.3\%$ while IC stays within 0.003 of the full method. The two stages barely touch the information metrics and act almost entirely on portfolio outcomes.

\textbf{The discovery components create the alpha itself.} Removing the searching strategy lowers IC by up to 0.002 relative to the discovery only variant and decreases the CUSTOM excess return from $-0.3\%$ to $-2.4\%$, and removing the dynamics costs up to 3.8 percentage points of excess return against the same control, the largest single decrease in the study. Unlike the downstream stages, these components influence information and portfolio metrics at once, so exploration quality cannot be repaired later in the pipeline.

\textbf{No single component explains the advantage.} The two partial methods in the first two parts still improve on the best agentic baseline in nearly every cell, with excess returns up to 2.6 percentage points higher, so the lead of AutoScientist-Quant is the sum of stage level and component level contributions rather than the effect of one source.

\subsection{Model Robustness Analysis}
\label{sec:model-robustness}

Table~\ref{tab:method-comparison} repeats the comparison on CSI 300 with the GLM backbone, covering the two agentic baselines, the discovery only variant, and the full method.

\textbf{The ranking holds up across the backbone change.} The full method again delivers the best information quality and the only consistent profits. IC reaches 0.036 with ALPHA158, above 0.032 for both baselines, ARR stays positive in all four settings, and seven of the eight baseline settings lose money after costs.

\textbf{The stage contributions are backbone independent.} Under GLM the full method outperforms the discovery only variant by 0.8 to 1.2 percentage points of excess return with a shallower MDD in every setting, matching the ablation trend found under GPT. GLM also lifts the absolute IC of nearly every method and setting, so the backbone sets the level of information quality while the framework sets the ordering.

\begin{table}[t!]
\centering
\setlength{\tabcolsep}{6pt}
\renewcommand{\arraystretch}{1.2}
\fontsize{8}{10}\selectfont
\resizebox{0.85\textwidth}{!}{%
\begin{tabular}{l c c c c c c c c}
\hline
\textbf{Alpha} & \textbf{IC} & \textbf{ICIR} & \textbf{RIC} & \textbf{RICIR} & \textbf{ARR} & \textbf{IR} & \textbf{MDD} & \textbf{CR} \\
\hline
\multicolumn{9}{c}{\textbf{AlphaAgent}} \\
\hline
CUSTOM    & \tcell{\phantom{-0.000}}{0.026}{\negv{-0.007}} & \tcell{\phantom{-0.000}}{0.148}{\negv{-0.052}} & \tcell{\phantom{-0.000}}{0.024}{\negv{-0.007}} & \tcell{\phantom{-0.000}}{0.139}{\negv{-0.053}} & \tcell{\phantom{-0.0\%}}{-3.1\%}{\negv{-3.7\%}} & \tcell{\phantom{-0.000}}{-0.463}{\negv{-0.554}} & \tcell{\phantom{-0.0\%}}{-15.3\%}{\negv{-6.7\%}} & \tcell{\phantom{-0.000}}{-0.191}{\negv{-0.262}} \\[\blockrowsep]
+ALPHA20  & \tcell{\phantom{-0.000}}{0.026}{\negv{-0.007}} & \tcell{\phantom{-0.000}}{0.156}{\negv{-0.054}} & \tcell{\phantom{-0.000}}{0.024}{\negv{-0.007}} & \tcell{\phantom{-0.000}}{0.147}{\negv{-0.055}} & \tcell{\phantom{-0.0\%}}{-1.1\%}{\negv{-2.2\%}} & \tcell{\phantom{-0.000}}{-0.159}{\negv{-0.319}} & \tcell{\phantom{-0.0\%}}{-14.5\%}{\negv{-4.6\%}} & \tcell{\phantom{-0.000}}{-0.034}{\negv{-0.175}} \\[\blockrowsep]
+ALPHA158 & \tcell{\phantom{-0.000}}{0.032}{\negv{-0.004}} & \tcell{\phantom{-0.000}}{0.199}{\negv{-0.025}} & \tcell{\phantom{-0.000}}{0.030}{\negv{-0.004}} & \tcell{\phantom{-0.000}}{0.187}{\negv{-0.030}} & \tcell{\phantom{-0.0\%}}{0.9\%}{\negv{-1.6\%}} & \tcell{\phantom{-0.000}}{0.141}{\negv{-0.243}} & \tcell{\phantom{-0.0\%}}{-11.0\%}{\negv{-2.4\%}} & \tcell{\phantom{-0.000}}{0.116}{\negv{-0.173}} \\[\blockrowsep]
+ALPHA360 & \tcell{\phantom{-0.000}}{0.028}{\negv{-0.007}} & \tcell{\phantom{-0.000}}{0.168}{\negv{-0.045}} & \tcell{\phantom{-0.000}}{0.026}{\negv{-0.007}} & \tcell{\phantom{-0.000}}{0.156}{\negv{-0.049}} & \tcell{\phantom{-0.0\%}}{-3.3\%}{\negv{-4.4\%}} & \tcell{\phantom{-0.000}}{-0.452}{\negv{-0.622}} & \tcell{\phantom{-0.0\%}}{-16.2\%}{\negv{-6.6\%}} & \tcell{\phantom{-0.000}}{-0.190}{\negv{-0.311}} \\
\hline
\multicolumn{9}{c}{\textbf{QuantaAlpha}} \\
\hline
CUSTOM    & \tcell{\phantom{-0.000}}{0.028}{\negv{-0.005}} & \tcell{\phantom{-0.000}}{0.174}{\negv{-0.026}} & \tcell{\phantom{-0.000}}{0.026}{\negv{-0.005}} & \tcell{\phantom{-0.000}}{0.165}{\negv{-0.027}} & \tcell{\phantom{-0.0\%}}{-3.6\%}{\negv{-4.2\%}} & \tcell{\phantom{-0.000}}{-0.566}{\negv{-0.656}} & \tcell{\phantom{-0.0\%}}{-15.1\%}{\negv{-6.5\%}} & \tcell{\phantom{-0.000}}{-0.222}{\negv{-0.294}} \\[\blockrowsep]
+ALPHA20  & \tcell{\phantom{-0.000}}{0.026}{\negv{-0.007}} & \tcell{\phantom{-0.000}}{0.159}{\negv{-0.050}} & \tcell{\phantom{-0.000}}{0.024}{\negv{-0.007}} & \tcell{\phantom{-0.000}}{0.150}{\negv{-0.052}} & \tcell{\phantom{-0.0\%}}{-2.0\%}{\negv{-3.0\%}} & \tcell{\phantom{-0.000}}{-0.307}{\negv{-0.467}} & \tcell{\phantom{-0.0\%}}{-15.6\%}{\negv{-5.7\%}} & \tcell{\phantom{-0.000}}{-0.061}{\negv{-0.203}} \\[\blockrowsep]
+ALPHA158 & \tcell{\phantom{-0.000}}{0.032}{\negv{-0.004}} & \tcell{\phantom{-0.000}}{0.198}{\negv{-0.026}} & \tcell{\phantom{-0.000}}{0.030}{\negv{-0.005}} & \tcell{\phantom{-0.000}}{0.186}{\negv{-0.030}} & \tcell{\phantom{-0.0\%}}{-1.2\%}{\negv{-3.8\%}} & \tcell{\phantom{-0.000}}{-0.178}{\negv{-0.562}} & \tcell{\phantom{-0.0\%}}{-11.7\%}{\negv{-3.1\%}} & \tcell{\phantom{-0.000}}{-0.095}{\negv{-0.385}} \\[\blockrowsep]
+ALPHA360 & \tcell{\phantom{-0.000}}{0.029}{\negv{-0.006}} & \tcell{\phantom{-0.000}}{0.175}{\negv{-0.037}} & \tcell{\phantom{-0.000}}{0.027}{\negv{-0.006}} & \tcell{\phantom{-0.000}}{0.164}{\negv{-0.041}} & \tcell{\phantom{-0.0\%}}{-2.2\%}{\negv{-3.3\%}} & \tcell{\phantom{-0.000}}{-0.297}{\negv{-0.467}} & \tcell{\phantom{-0.0\%}}{-13.9\%}{\negv{-4.2\%}} & \tcell{\phantom{-0.000}}{-0.155}{\negv{-0.276}} \\
\hline
\multicolumn{9}{c}{\textbf{AutoScientist-Quant (w/o Alpha Filtering or Model Search) = Alpha Discovery}} \\
\hline
CUSTOM    & \tcell{\posv{+0.002}}{0.031}{\negv{-0.002}} & \tcell{\posv{+0.010}}{0.184}{\negv{-0.016}} & \tcell{\posv{+0.003}}{0.029}{\negv{-0.002}} & \tcell{\posv{+0.013}}{0.177}{\negv{-0.015}} & \tcell{\posv{+2.7\%}}{-0.4\%}{\negv{-1.0\%}} & \tcell{\posv{+0.399}}{-0.064}{\negv{-0.155}} & \tcell{\posv{+5.0\%}}{-10.1\%}{\negv{-1.5\%}} & \tcell{\posv{+0.153}}{-0.038}{\negv{-0.109}} \\
+ALPHA20  & \tcell{\posv{+0.005}}{0.032}{\negv{-0.002}} & \tcell{\posv{+0.033}}{0.193}{\negv{-0.017}} & \tcell{\posv{+0.006}}{0.030}{\negv{-0.001}} & \tcell{\posv{+0.035}}{0.185}{\negv{-0.017}} & \tcell{\posv{+1.2\%}}{0.1\%}{\negv{-1.0\%}} & \tcell{\posv{+0.166}}{0.006}{\negv{-0.154}} & \tcell{\posv{+4.2\%}}{-10.3\%}{\negv{-0.4\%}} & \tcell{\posv{+0.098}}{0.064}{\negv{-0.077}} \\
+ALPHA158 & \tcell{\posv{+0.002}}{0.034}{\negv{-0.002}} & \tcell{\posv{+0.008}}{0.206}{\negv{-0.018}} & \tcell{\posv{+0.002}}{0.032}{\negv{-0.002}} & \tcell{\posv{+0.009}}{0.196}{\negv{-0.021}} & \tcell{\posv{+0.5\%}}{1.4\%}{\negv{-1.2\%}} & \tcell{\posv{+0.061}}{0.202}{\negv{-0.182}} & \tcell{\posv{+1.6\%}}{-9.5\%}{\negv{-0.8\%}} & \tcell{\posv{+0.024}}{0.140}{\negv{-0.149}} \\
+ALPHA360 & \tcell{\posv{+0.004}}{0.033}{\negv{-0.002}} & \tcell{\posv{+0.021}}{0.196}{\negv{-0.016}} & \tcell{\posv{+0.004}}{0.031}{\negv{-0.002}} & \tcell{\posv{+0.023}}{0.188}{\negv{-0.018}} & \tcell{\posv{+2.4\%}}{0.3\%}{\negv{-0.8\%}} & \tcell{\posv{+0.341}}{0.044}{\negv{-0.126}} & \tcell{\posv{+3.8\%}}{-10.1\%}{\negv{-0.5\%}} & \tcell{\posv{+0.187}}{0.032}{\negv{-0.089}} \\
\hline
\multicolumn{9}{c}{\textbf{AutoScientist-Quant = Alpha Discovery + Alpha Filtering + Model Search}} \\
\hline
CUSTOM    & \tcell{\posv{+0.005}}{0.033}{\phantom{-0.000}} & \tcell{\posv{+0.026}}{0.200}{\phantom{-0.000}} & \tcell{\posv{+0.005}}{0.031}{\phantom{-0.000}} & \tcell{\posv{+0.027}}{0.192}{\phantom{-0.000}} & \tcell{\posv{+3.7\%}}{0.6\%}{\phantom{-0.0\%}} & \tcell{\posv{+0.554}}{0.091}{\phantom{-0.000}} & \tcell{\posv{+6.5\%}}{-8.6\%}{\phantom{-0.0\%}} & \tcell{\posv{+0.262}}{0.071}{\phantom{-0.000}} \\[\blockrowsep]
+ALPHA20  & \tcell{\posv{+0.007}}{0.033}{\phantom{-0.000}} & \tcell{\posv{+0.050}}{0.209}{\phantom{-0.000}} & \tcell{\posv{+0.007}}{0.031}{\phantom{-0.000}} & \tcell{\posv{+0.052}}{0.202}{\phantom{-0.000}} & \tcell{\posv{+2.2\%}}{1.1\%}{\phantom{-0.0\%}} & \tcell{\posv{+0.319}}{0.160}{\phantom{-0.000}} & \tcell{\posv{+4.6\%}}{-9.9\%}{\phantom{-0.0\%}} & \tcell{\posv{+0.175}}{0.142}{\phantom{-0.000}} \\[\blockrowsep]
+ALPHA158 & \tcell{\posv{+0.004}}{0.036}{\phantom{-0.000}} & \tcell{\posv{+0.025}}{0.224}{\phantom{-0.000}} & \tcell{\posv{+0.004}}{0.035}{\phantom{-0.000}} & \tcell{\posv{+0.030}}{0.216}{\phantom{-0.000}} & \tcell{\posv{+1.6\%}}{2.5\%}{\phantom{-0.0\%}} & \tcell{\posv{+0.243}}{0.384}{\phantom{-0.000}} & \tcell{\posv{+2.4\%}}{-8.6\%}{\phantom{-0.0\%}} & \tcell{\posv{+0.173}}{0.289}{\phantom{-0.000}} \\[\blockrowsep]
+ALPHA360 & \tcell{\posv{+0.006}}{0.035}{\phantom{-0.000}} & \tcell{\posv{+0.037}}{0.212}{\phantom{-0.000}} & \tcell{\posv{+0.006}}{0.033}{\phantom{-0.000}} & \tcell{\posv{+0.041}}{0.205}{\phantom{-0.000}} & \tcell{\posv{+3.3\%}}{1.1\%}{\phantom{-0.0\%}} & \tcell{\posv{+0.467}}{0.170}{\phantom{-0.000}} & \tcell{\posv{+4.2\%}}{-9.6\%}{\phantom{-0.0\%}} & \tcell{\posv{+0.276}}{0.121}{\phantom{-0.000}} \\
\hline
\end{tabular}%
}
\caption{\textbf{Model robustness on CSI 300 under GLM.} The center value in each cell is the absolute metric, the top right corner is the gain over the best baseline value per metric, and the bottom right corner is the gap to our full method, both computed from unrounded values. \textcolor{darkgreen}{Green} marks positive values and \textcolor{darkred}{red} negative.}
\label{tab:method-comparison}
\vspace{-2em}
\end{table}

\begin{table}[t!]
\centering
\setlength{\tabcolsep}{7pt}
\fontsize{6}{8}\selectfont
\begin{tabular}{l l c c c c c c c c}
\hline
\textbf{Backbone} & \textbf{Method} & \textbf{IC} & \textbf{ICIR} & \textbf{RIC} & \textbf{RICIR} & \textbf{ARR} & \textbf{IR} & \textbf{MDD} & \textbf{CR} \\
\hline
\multicolumn{10}{c}{\textbf{\textit{CSI 500}}} \\
\hline
\multirow{3}{*}{GPT}
 & AlphaAgent          & 0.047 & 0.312 & 0.044 & 0.290 & -0.7\% & -0.076 & -19.4\% & -0.035 \\
 & QuantaAlpha         & 0.047 & 0.322 & 0.043 & 0.299 & 0.2\%  & 0.022  & -17.7\% & 0.009 \\
 & \textbf{AutoScientist-Quant} & \textbf{0.049} & \textbf{0.328} & \textbf{0.045} & \textbf{0.303} & \textbf{0.5\%} & \textbf{0.062} & \textbf{-17.2\%} & \textbf{0.032} \\
\hline
\multirow{3}{*}{GLM}
 & AlphaAgent          & 0.047 & 0.328 & 0.043 & 0.302 & -0.1\% & -0.016 & -17.3\% & -0.008 \\
 & QuantaAlpha         & \textbf{0.048} & 0.331 & 0.044 & 0.305 & -1.3\% & -0.143 & -16.7\% & -0.077 \\
 & \textbf{AutoScientist-Quant} & 0.048 & \textbf{0.341} & \textbf{0.044} & \textbf{0.315} & \textbf{-0.1\%} & \textbf{-0.014} & \textbf{-15.2\%} & \textbf{-0.007} \\
\hline
\multicolumn{10}{c}{\textbf{\textit{CSI 800}}} \\
\hline
\multirow{3}{*}{GPT}
 & AlphaAgent          & 0.043 & 0.297 & 0.039 & 0.275 & 0.8\%  & 0.088  & -19.3\% & 0.063 \\
 & QuantaAlpha         & 0.043 & 0.313 & 0.040 & 0.291 & 0.2\%  & 0.023  & -19.1\% & 0.012 \\
 & \textbf{AutoScientist-Quant} & \textbf{0.045} & \textbf{0.318} & \textbf{0.041} & \textbf{0.298} & \textbf{2.4\%} & \textbf{0.261} & \textbf{-17.9\%} & \textbf{0.135} \\
\hline
\multirow{3}{*}{GLM}
 & AlphaAgent          & 0.044 & 0.309 & 0.040 & 0.289 & -0.2\% & -0.018 & -19.8\% & -0.007 \\
 & QuantaAlpha         & 0.044 & 0.307 & 0.041 & 0.285 & 0.9\%  & 0.094  & -20.2\% & 0.055 \\
 & \textbf{AutoScientist-Quant} & \textbf{0.046} & \textbf{0.317} & \textbf{0.042} & \textbf{0.292} & \textbf{1.3\%} & \textbf{0.133} & \textbf{-18.5\%} & \textbf{0.074} \\
\hline
\multicolumn{10}{c}{\textbf{\textit{CSI 1000}}} \\
\hline
\multirow{3}{*}{GPT}
 & AlphaAgent          & 0.065 & 0.469 & 0.058 & 0.417 & 6.4\%  & 0.654  & -15.2\% & 0.429 \\
 & QuantaAlpha         & \textbf{0.067} & \textbf{0.485} & 0.060 & 0.431 & 4.8\%  & 0.489  & -13.0\% & 0.371 \\
 & \textbf{AutoScientist-Quant} & 0.066 & 0.481 & \textbf{0.060} & \textbf{0.437} & \textbf{7.1\%} & \textbf{0.735} & \textbf{-12.9\%} & \textbf{0.591} \\
\hline
\multirow{3}{*}{GLM}
 & AlphaAgent          & 0.065 & 0.486 & 0.058 & 0.437 & 6.5\%  & 0.687  & -13.3\% & 0.533 \\
 & QuantaAlpha         & 0.066 & 0.485 & 0.058 & 0.427 & 6.4\%  & 0.671  & -13.5\% & 0.471 \\
 & \textbf{AutoScientist-Quant} & \textbf{0.068} & \textbf{0.496} & \textbf{0.061} & \textbf{0.444} & \textbf{6.9\%} & \textbf{0.711} & \textbf{-11.6\%} & \textbf{0.600} \\
\hline
\end{tabular}
\caption{\textbf{Market robustness on CSI 500, CSI 800, and CSI 1000.} The three agentic methods run under both the GPT and the GLM backbone, with alphas discovered within each universe. \textbf{Bold} marks the best method in each setting, based on unrounded original values, so equal displayed values can differ in bolding.}
\label{tab:ablation-market}
\vspace{-1em}
\end{table}

\subsection{Market Robustness Analysis}
\label{sec:market-robustness}

Table~\ref{tab:ablation-market} evaluates the three agentic methods on CSI 500, CSI 800, and CSI 1000 under both backbones, and Table~\ref{tab:us-markets} extends the comparison to S\&P 500 and NASDAQ 100 under the GLM backbone. Alphas are discovered within each universe rather than being transferred, so the analysis tests whether the framework itself, not one particular alpha set, generalizes.

\begin{table}[t!]
\centering
\setlength{\tabcolsep}{7pt}
\fontsize{6}{8}\selectfont
\begin{tabular}{l l c c c c c c c c}
\hline
\textbf{Method} & \textbf{Alpha} & \textbf{IC} & \textbf{ICIR} & \textbf{RIC} & \textbf{RICIR} & \textbf{ARR} & \textbf{IR} & \textbf{MDD} & \textbf{CR} \\
\hline
\multicolumn{10}{c}{\textbf{\textit{S\&P 500}}} \\
\hline
\multirow{4}{*}{AlphaAgent}
 & CUSTOM    & 0.006 & 0.043 & 0.005 & 0.040 & 10.4\% & 0.687 & -21.1\% & 0.476 \\
 & +ALPHA20  & 0.005 & 0.035 & 0.005 & 0.036 & 9.8\%  & 0.632 & -24.2\% & 0.399 \\
 & +ALPHA158 & 0.004 & 0.028 & 0.002 & 0.019 & 12.4\% & 0.860 & -20.5\% & 0.606 \\
 & +ALPHA360 & 0.005 & 0.035 & 0.004 & 0.030 & 10.3\% & 0.674 & -26.5\% & 0.386 \\
\hline
\multirow{4}{*}{QuantaAlpha}
 & CUSTOM    & 0.007 & 0.051 & 0.007 & 0.053 & 12.3\% & 0.816 & -22.5\% & 0.538 \\
 & +ALPHA20  & 0.006 & 0.043 & 0.006 & 0.046 & 13.0\% & 0.857 & \textbf{-21.4\%} & 0.615 \\
 & +ALPHA158 & 0.005 & 0.037 & 0.004 & 0.033 & 10.7\% & 0.758 & -21.7\% & 0.494 \\
 & +ALPHA360 & 0.006 & 0.043 & 0.006 & 0.041 & 10.7\% & 0.714 & -26.4\% & 0.405 \\
\hline
\multirow{4}{*}{AutoScientist-Quant}
 & CUSTOM    & \textbf{0.009} & \textbf{0.073} & \textbf{0.009} & \textbf{0.071} & \textbf{15.4\%} & \textbf{1.021} & \textbf{-19.8\%} & \textbf{0.795} \\
 & +ALPHA20  & \textbf{0.011} & \textbf{0.080} & \textbf{0.010} & \textbf{0.073} & \textbf{15.2\%} & \textbf{0.982} & -22.3\% & \textbf{0.681} \\
 & +ALPHA158 & \textbf{0.007} & \textbf{0.053} & \textbf{0.006} & \textbf{0.048} & \textbf{12.8\%} & \textbf{0.905} & \textbf{-19.8\%} & \textbf{0.648} \\
 & +ALPHA360 & \textbf{0.008} & \textbf{0.059} & \textbf{0.007} & \textbf{0.051} & \textbf{11.5\%} & \textbf{0.759} & \textbf{-25.3\%} & \textbf{0.453} \\
\hline
\multicolumn{10}{c}{\textbf{\textit{NASDAQ 100}}} \\
\hline
\multirow{4}{*}{AlphaAgent}
 & CUSTOM    & 0.009 & 0.049 & 0.010 & 0.051 & 20.1\% & 1.083 & \textbf{-20.9\%} & 0.970 \\
 & +ALPHA20  & 0.006 & 0.032 & 0.006 & 0.036 & 13.7\% & 0.781 & -20.8\% & 0.695 \\
 & +ALPHA158 & 0.003 & 0.014 & 0.002 & 0.012 & 12.1\% & 0.749 & -19.6\% & 0.609 \\
 & +ALPHA360 & 0.003 & 0.017 & 0.004 & 0.022 & 13.1\% & 0.799 & -19.7\% & 0.662 \\
\hline
\multirow{4}{*}{QuantaAlpha}
 & CUSTOM    & 0.010 & 0.053 & 0.010 & 0.052 & 22.3\% & 1.041 & -25.0\% & 0.877 \\
 & +ALPHA20  & 0.001 & 0.005 & 0.002 & 0.012 & 17.8\% & 1.028 & -18.6\% & 0.953 \\
 & +ALPHA158 & 0.001 & 0.004 & 0.000 & 0.002 & 13.3\% & 0.883 & \textbf{-17.9\%} & 0.749 \\
 & +ALPHA360 & 0.000 & 0.001 & 0.000 & 0.002 & 12.6\% & 0.818 & -19.2\% & 0.648 \\
\hline
\multirow{4}{*}{AutoScientist-Quant}
 & CUSTOM    & \textbf{0.018} & \textbf{0.091} & \textbf{0.018} & \textbf{0.091} & \textbf{28.9\%} & \textbf{1.297} & -22.5\% & \textbf{1.289} \\
 & +ALPHA20  & \textbf{0.011} & \textbf{0.059} & \textbf{0.011} & \textbf{0.060} & \textbf{24.6\%} & \textbf{1.280} & \textbf{-18.5\%} & \textbf{1.323} \\
 & +ALPHA158 & \textbf{0.007} & \textbf{0.034} & \textbf{0.007} & \textbf{0.035} & \textbf{18.3\%} & \textbf{1.053} & -18.7\% & \textbf{1.010} \\
 & +ALPHA360 & \textbf{0.006} & \textbf{0.032} & \textbf{0.007} & \textbf{0.035} & \textbf{16.3\%} & \textbf{0.949} & \textbf{-19.0\%} & \textbf{0.887} \\
\hline
\end{tabular}
\caption{\textbf{Market robustness on S\&P 500 and NASDAQ 100 under GLM.} Each cell is the absolute metric computed without subtracting the index return, since US equity strategies are commonly evaluated this way \citep{sharpe1998sharpe} rather than by the index excess used on the CSI universes. \textbf{Bold} marks the best method in each setting, as in Table~\ref{tab:ablation-market}.}
\vspace{-2em}
\label{tab:us-markets}
\end{table}

\textbf{The advantage holds in every market and under both backbones.} AutoScientist-Quant delivers the best or equal best excess return in all six settings and the best value on nearly all other metrics. The exceptions are individual information metrics in two settings, such as the IC of QuantaAlpha on CSI 1000 under GPT at 0.067 against our 0.066, where our portfolio metrics remain the strongest.

\textbf{Gains concentrate where pricing is least efficient.} For all methods IC rises from between 0.043 and 0.049 on CSI 500 and CSI 800 to between 0.065 and 0.068 on CSI 1000, and our excess return climbs from 0.5\% on CSI 500 to 7.1\% on CSI 1000 under GPT and from $-0.1\%$ to 6.9\% under GLM. This matches the common view that small caps have more remaining alpha.

\textbf{The framework holds up in hard markets and pushes in easy ones.} On the efficiently priced large and mid cap universes the baselines often lose after costs, such as AlphaAgent at $-0.7\%$ on CSI 500 under GPT, while AutoScientist-Quant stays ahead with the shallowest MDD. On CSI 1000 it extends the best baseline result from 6.4\% to 7.1\% under GPT and reaches a CR of 0.600 under GLM.

\vspace{-0.5em}
\subsection{Case Study}
\label{sec:case-study}
\vspace{-0.5em}

\begin{figure}[t!]
\centering
\includegraphics[width=\textwidth]{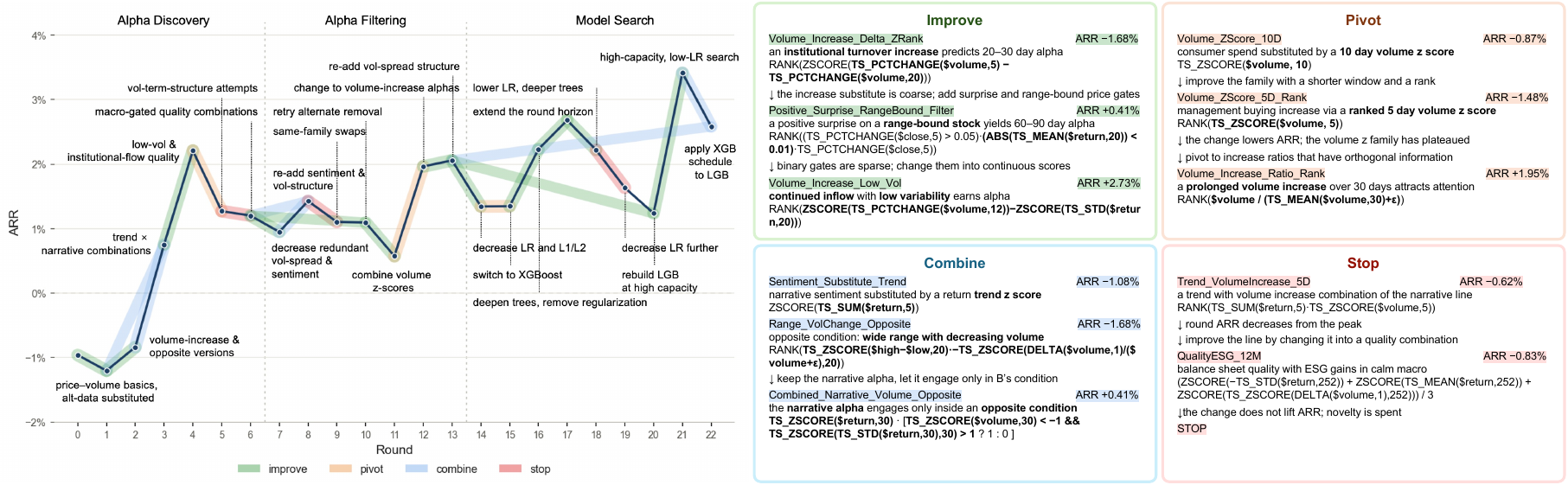}
\vspace{-2em}
\caption{\textbf{Example of one search.} Left: test window ARR of the +ALPHA20 deployment after each round of the CSI 300 search under GPT; the search itself sees only the feedback window. Colors mark the option of each move. Right: one example per option with its alpha names, ARR, and the reasoning and reflection of the model.}
\label{fig:case-study}
\vspace{-1em}
\end{figure}

Figure~\ref{fig:case-study} traces the searches behind Table~\ref{tab:alpha-results}; four behaviors stand out.

\textbf{Test performance recovers from dips the search cannot see.} Filtering touches its stage low one move before the swap that lifts the subset to its stage best, and model search leaves an XGBoost branch, restarts from the root, loses further ground, and reaches the stage best on the next move. The pool of stored parents lets the search return to a promising branch after a period of losses. The curve reports the test window while every move is chosen on the feedback window alone, so these dips and recoveries are invisible to the search and appear only in hindsight.

\textbf{A few redundant members move ARR more than library size.} The selected subset keeps 89 of 91 alphas yet single swaps move test ARR by more than a percentage point. Adding two same family volume z score alphas brings the subset to its stage low, while replacing two relative volume members with two volume increase members lifts it to the stage best. The margin lies in redundancy control rather than raw library size.

\textbf{Revision accumulates along histories.} \textsc{Improve} rewrites a turnover increase substitute into an earnings surprise gate on flat priced stocks, then changes the binary gates into continuous scores, two revisions of one history rather than fresh samples; the rounds that add its three generations backtest at $-1.7\%$, $+0.4\%$, and $+2.7\%$. \textsc{Combine} keeps the trend term of a sentiment substitute parent and then uses it only inside the opposite condition of a range and volume parent; the sentiment parent never enters the final library although its children do.

\textbf{Stopping is evidential, not scheduled.} Revising a trend and volume combination at $-0.6\%$ into a one year quality combination lowers ARR to $-0.8\%$, and the stage stops with two alphas produced in its final round; filtering spends fourteen rounds confirming that no edit outperforms the selected subset, then rolls back to it. The budget rule of Eq.~\ref{eq:budget} makes this policy. Capacity follows marginal evidence and stops once it is not obvious.

\vspace{-0.5em}
\section{Conclusions}
\label{Conclusions}
\vspace{-0.5em}

AutoScientist-Quant runs alpha discovery, alpha filtering, and model search under one controller that regards option choice and spend as decisions made during the run. On CSI universes it leads on nearly every metric in every setting, under two backbones and across four universes, and the ablations and case study explain the reasons. Exploration creates the information, downstream stages turn it into returns, and the search wins by recovering from dips it cannot see, controlling redundancy, and stopping on evidence.

In the future, the same search framework can be used in other domains such as algorithm development \citep{li2025a, NEURIPS2025_77f33e8b, 11462836, romera2024mathematical}, reasoning \citep{yao2023tree, shinn2023reflexion, li-etal-2025-reasongraph, li-etal-2025-500xcompressor}, and evaluation \citep{zhou2026general}.

\bibliography{custom}
\bibliographystyle{iclr2027_conference}

\end{document}